\documentclass[conference]{IEEEtran}
\pdfoutput=1
\usepackage{amsfonts}
\IEEEoverridecommandlockouts
\usepackage{cite}
\usepackage[linesnumbered,ruled,vlined]{algorithm2e}
\usepackage{amsmath,amssymb,amsfonts}
\usepackage{graphicx}
\usepackage{epstopdf}
\usepackage{textcomp}
\usepackage{xcolor}
\usepackage[colorlinks, linkcolor=blue]{hyperref}
\usepackage{caption}
\usepackage{multirow}
\usepackage{subfigure}
\def\BibTeX{{\rm B\kern-.05em{\sc i\kern-.025em b}\kern-.08em
    T\kern-.1667em\lower.7ex\hbox{E}\kern-.125emX}}

\begin{document}

\title{FedHe: Heterogeneous Models and Communication-Efficient Federated Learning\\
{\footnotesize \textsuperscript{}}
\thanks{Accepted by The 17th International Conference on Mobility, Sensing and Networking (MSN 2021)}
}

\newcommand{\encom}[1]{\footnote{{\bf \color{purple}Edith: #1}}}

\author{\IEEEauthorblockN{Yun Hin Chan}
\IEEEauthorblockA{\textit{Department of Electrical and Electronic Engineering} \\
\textit{The University of Hong Kong}\\
yhchan@eee.hku.hk}

\and
\IEEEauthorblockN{Edith C.H. Ngai}
\IEEEauthorblockA{\textit{Department of Electrical and Electronic Engineering} \\
\textit{The University of Hong Kong}\\
chngai@eee.hku.hk}
}

\maketitle

\begin{abstract}
Federated learning (FL) is able to manage edge devices to cooperatively train a model while maintaining the training data local and private. One common assumption in FL is that all edge devices share the same machine learning model in training, for example, identical neural network architecture.
 However, the computation and store capability of different devices may not be the same. Moreover, reducing communication overheads can improve the training efficiency though it is still a challenging problem in FL. In this paper, we propose a novel FL method, called FedHe, inspired by knowledge distillation, which can train  heterogeneous models and support asynchronous training processes with significantly reduced communication overheads. Our analysis and experimental results demonstrate that the performance of our proposed method is better than the state-of-the-art algorithms in terms of communication overheads and model accuracy.
\end{abstract}


\section{Introduction}
Massive data is created with the proliferation of edge devices. Handling large amount of data becomes a challenging problem in this information era. Distributed machine learning \cite{verbraeken2020survey} is proposed to coordinate massive data management in the data centers, which is turning into a crucial technique. However, accessing and processing data with distributed machine learning algorithms at remote servers has raised data privacy concerns from the users.
\textbf{Federated learning} (FL) has been proposed as a machine learning paradigm that aims to perform machine learning on massive and distributed data without invading data privacy. More concisely, FL coordinates clients and servers to train a shared global model, while keeping the data locally at the edge devices. The general FL architecture was first proposed in \cite{mcmahan2017communication}, which is composed of servers and clients. The work has attracted a lot of attention and led to many follow-up studies \cite{li2020federated}\cite{bonawitz2019towards}\cite{konevcny2016federated}\cite{chai2020tifl}.

Although many researchers focus on FL and there are many applications such as predicting human activities \cite{chen2019communication} and learning sentiment\cite{smith2017federated}, it still has many practical challenges to be solved \cite{kairouz2019advances}. One of the most important challenges comes from system heterogeneity, which causes many issues in FL. System heterogeneity usually refers to various computation and bandwidth resources among different participating devices. It would lead to a common but undesirable situation that incompetent devices drag the convergence speed of the server model. These devices, called stragglers, are not able to complete their training when the other efficient participants finish their model training processes. However, many FL algorithms \cite{mcmahan2017communication}\cite{sattler2019robust}\cite{li2018federated} are not designed to handle this practical problem.

Some studies focus on how to design an asynchronous training scheme \cite{xie2019asynchronous}\cite{chen2020asynchronous}\cite{chai2020fedat} to alleviate this problem. This is a straightforward way to deal with the problem caused by stragglers.
Different from the above works, we take a new approach to handle system heterogeneity and provide a more general and practical solution by supporting heterogeneous model architectures for devices with different capabilities. For example, the model architectures of neural networks could be chosen individually, depending on the computation power and bandwidth resources of the participating devices. When the devices have chosen suitable model structures, they are able to complete the training process faster and potentially simultaneously.

In traditional FL  \cite{mcmahan2017communication}, model weights or gradients are aggregated in the server and shared with the local devices. Although local data are kept private, exchanging model parameters between the clients and the serve occupies a lot of bandwidth. Moreover, the model architectures of the clients and the server have to be identical in order to aggregate the model in common FL methods. A few research studies have attempted to support heterogeneous models in FL \cite{li2019fedmd}\cite{li2021fedh2l}. However, their approaches require a public dataset in the server, which is comprised of a portion of each private dataset or an existing dataset. Obtaining part of the local private dataset from clients may violate user privacy. It is also difficult to have an existing dataset which is similar to the private datasets. Therefore, how to transfer knowledge between devices with heterogeneous model architectures still remains to be a research challenge.

To solve the above problem, we propose a novel method called Heterogeneous models and Efficient Federated learning (\textbf{FedHe}), inspired by knowledge distillation (KD) \cite{hinton2015distilling}. FedHe takes an asynchronous FL approach and supports training of heterogeneous models with small communication overheads.
Knowledge distillation is a model compression method, which a heavy teacher model distills knowledge and transmits it to a light student model. In our method, we conduct a training process using this knowledge, generated by the clients and processed in the server, to replace the exchanges and updates of weights in the original FL methods. In our approach, participants can share knowledge with each other successfully without deploying a public dataset. Our method can run in an asynchronous mode, which is easier to deploy in real FL environment and preserve user privacy. Our experimental results show that our algorithm converges in an asynchronous setting with satisfactory model accuracy and significantly reduced communication overheads.

The rest paper is organized as follows. The related work about federated learning is presented in Section 2. The problem formulation is described in Section 3. The details of our proposed method in FedHe is presented in Section 4. Section 5 shows the experimental results and compares FedHe with the existing work. We conclude this paper in Section 6.

\section{Related Work}
\subsection{Federated learning}
Federated learning (FL), was first proposed by Google \cite{mcmahan2017communication} in 2017. It provides a novel method to organize computing among edge devices. It is a synchronous algorithm that can compute model weights from randomly selected clients iteratively. FedProx \cite{li2018federated} manages the training process synchronously under non-i.i.d. data setting. However, many of the FL schemes are synchronous algorithms.
FedAsyn \cite{xie2019asynchronous} makes an asynchronous training scheme with coordinators and schedulers, and uses a weighted average to update the global model. Although our work does not focus on asynchronous FL, our proposed method, FedHe, can work in an asynchronous fashion, which makes it more practical and easier to be deployed in the real world.

\subsection{Knowledge distillation}
Hinton et al. proposed knowledge distillation for neural networks in  \cite{hinton2015distilling}. The initial goal of this method is to compress heavy neural networks. A heavy teacher model transmits its knowledge, called logits, to a light student model, while the student model performs training based on the same dataset and logits. Federated Distillation \cite{seo2020federated} discussed how to use logits in the homogeneous models FL. Kim et al. \cite{kim2018paraphrasing} used a paraphraser as a teacher network and a translator as a student model to communicate the latent knowledge, called ``factors", with each other. SemCKD \cite{chen2021cross} applied attention allocation to match intermediate knowledge from teacher layers to the appropriate student layers, which is a cross-layer distillation method. In this work, we proposes FedHe, inspired by the basic method in knowledge distillation \cite{hinton2015distilling}, to enable knowledge sharing among various clients in FL efficiently.

\subsection{Heterogeneous models}
Supporting heterogeneous models in FL environment is a big challenge. FedMD \cite{li2019fedmd} introduced logits from a large public dataset. The clients can learn the data features from the logits and their private datasets. FedML \cite{shen2020federated} applied latent information from homogeneous models to train heterogeneous models. FedH2L \cite{li2021fedh2l} adopted logits from a public dataset consisted of private data instances from different clients. It keeps the optimized directions from the public and private datasets the same. In contrast, we do not need a public dataset in our method. The logits, capturing knowledge from private datasets, are collected from the training processes of clients, but they do not invade user privacy. Our experiments show that FedHe can support FL with homogeneous and heterogeneous models, since the exchange of logits depends only on the number of output classes and is generic to different model architectures.

\section{Problem Formulation}
In this section, we show the general form of the FL problem and introduce the heterogeneous FL problem. For clarity, we list symbols frequently used in this paper in Table \ref{list_of_symbols}.

\subsection{General federated learning}
We consider $K$ devices (clients) in the FL framework. Each device $k$ has a private local dataset $D_k$, where $k\in\{1,...,K\}$. We denote the parameters of a training model by $w$. We formulate the minimization objective function for each device as follow,
\begin{equation}
\label{FL_local_objective}
\begin{aligned}
 \min&\ & L_k(w)=\frac{1}{|D_k|}\sum_{i=1}^{|D_k|}{l(f(x_i;w), y_i)},
\end{aligned}
\end{equation}
where $|D_k|$ is the number of instances in $D_k$, $(x_i, y_i)\in D_k$, and $l$ is a loss function for computing the loss of the predicted output $f(x_i;w)$ on instance $(x_i, y_i)$ with model parameters $w$. Let $N=\sum_{i=1}^K |D_k|$, we have a global convex optimization problem,
\begin{equation}
\label{FL_global_objective}
\begin{aligned}
 \min_w&\ & L(w)=\sum_{k=1}^K\frac{|D_k|}{N}{L_k(w)}.
\end{aligned}
\end{equation}
When we aggregate the loss from different devices, we multiply $L_k(w)$ by $|D_k|$ to consider the influence from the size of data in $k$. If a client has a large dataset, it is deserved to pay more attention to its loss. The most common solution is to compute the optimal parameters $w$ by gradient descent methods synchronously with a neural network model, such as in FedAvg \cite{mcmahan2017communication}.

\subsection{Heterogeneous federated learning}
Due to system heterogeneity, some clients in FL may have less memory and computation capability.
It may be difficult to apply the same machine learning model architecture to all the clients. Our idea is to support FL with heterogeneous models in the clients, so that the clients can learn models with appropriate size and computation complexity while still can share their knowledge.

We formulate the heterogeneous FL problem as follows. The model parameters of client $k$ is denoted by $w_k$. The objective function for client $k$ is similar to that in Eq. (\ref{FL_local_objective}). The model parameters in Eq. (\ref{FL_local_objective}), $w$, are substituted by $w_k$. The global objective function becomes
\begin{equation}
\label{Hetero_FL_global_objective}
\begin{aligned}
 \min_{w_1, w_2, ..., w_K}&\ & \sum_{k=1}^K\frac{|D_k|}{N}{L_k(w_k)}.
\end{aligned}
\end{equation}
The clients can have different model architectures, so the function of the server in this problem is not to aggregate the parameters like in Eq.  (\ref{FL_global_objective}). Instead, its objective is to train heterogeneous models for different clients. We present our method and explain how to optimize this objective function in the next section.
\begin{table}
  \centering
  \begin{tabular}{ll}
    \hline
    Symbol & Description \\
    \hline
    $D_k$ & A dataset on client $k$ \\
    $|D_k|$ & The size of $D_k$ \\
    $N$ & The size of datasets from all clients \\
    $K$ & The size of clients \\
    $w$ & The weights of global model \\
    $w_k$ & The weights of a model from the client $k$\\
    $L_k$ & The objective function of the client $k$ \\
    $L$ & The global objective function \\
    $p_i$ & The logit for the instance $(x_i, y_i)$ \\
    $f(x;w)$ & The neural network model with weights $w$ and inputs $x$ \\
    $V_{y_i}$ & The number of logits belong to $y_i$ \\
    $softmax(x)$ & The softmax layer with inputs $x$ \\
    $OneHot(x)$ & The One Hot function with inputs $x$ \\
    \hline
  \end{tabular}
  \caption{List of symbols}\label{list_of_symbols}
\end{table}
\section{Our Proposed Method}
\begin{figure*}[htbp]
\centering
    \subfigure{
    \begin{minipage}{12cm}
        \centering
        \includegraphics[width=12cm]{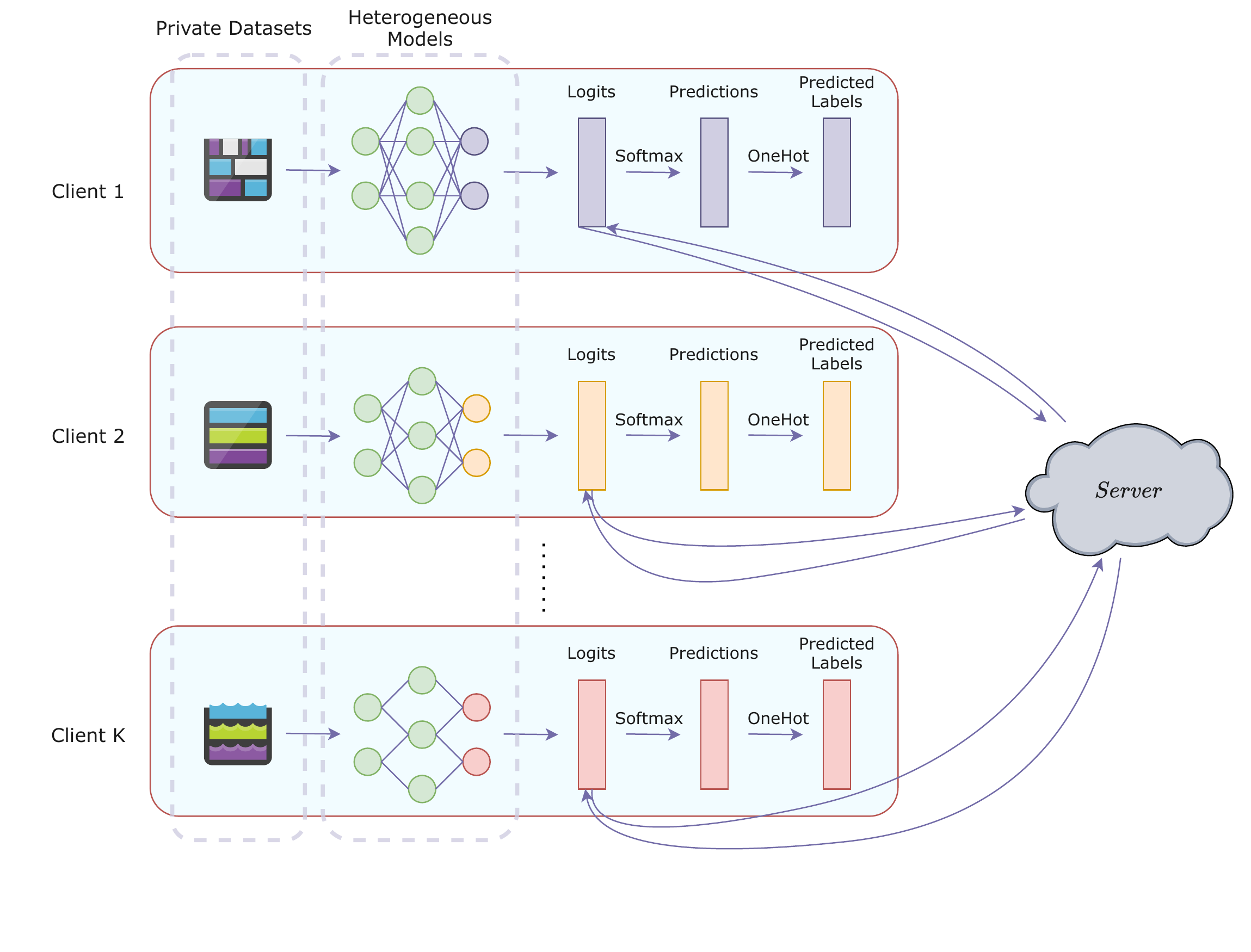}
        \caption{The system architecture of FedHe. In our method, clients have different private dataset and their models are heterogeneous. In the training process, clients not only train on the private dataset, they also train on logits came from the server. Clients collect the logits from private datasets and send them to the server. The server processes these logits and transmits back to clients.}
        \label{fig_FedHe}
    \end{minipage}
    }
\end{figure*}

To solve the heterogeneous FL problem, we propose a novel method called \textbf{FedHe}, enlightened by KD. We apply logits, considered as knowledge from a teacher model in KD, to train heterogeneous models in different clients. FedHe can train clients with heterogeneous models asynchronously with small communication overheads in the network.  Fig.~\ref{fig_FedHe} shows the system architecture of FedHe. We describe our method in three parts, including training on private data, collection of logits, and training on logits in the following.

\subsection{Training on private data}
In this section, we describe how clients update their local models using their own private datasets. This process is a conventional supervised learning scheme, i.e., the client $k$ trains its local model $w_k$ based on its randomly selected private dataset $(x_i, y_i)\in \Tilde{D_k}$. The objective function of this optimization process is shown as,
\begin{equation}
\label{local_private_data_objective}
\begin{aligned}
\min_{w_k}&\ & L_{private}=\frac{1}{|\Tilde{D_k}|}\sum_{i=1}^{|\Tilde{D_k}|}{l_{private}(\Tilde{y_i}, y_i)},
\end{aligned}
\end{equation}
where $l_{private}$ is a cross-entropy function for input sample $x_i$ with label $y_i$ and predicted label $\Tilde{y_i}$ from the neural network. In this training process, the client randomly selects a batch of instances from its private dataset to train the model in each round.

\subsection{Collection of logits}
In our method, logits are defined as inputs of the softmax layer in the neural network, denoted by $p_i$, for instance $(x_i, y_i)$. The relationship between logit $p_i$ and $\Tilde{y_i}$ is described as,
\begin{equation}
\label{1_logit_output_relationship}
p_i = f(x_i;w_k)[-2],
\end{equation}
\begin{equation}
\label{2_logit_output_relationship}
\begin{aligned}
\Tilde{y_i} &=& OneHot(softmax(p_i)) \\
&=& OneHot(f(x_i;w_k)[-1]),
\end{aligned}
\end{equation}
where $f(x_i;w_k)[-2]$ is an output of the second last layer of the model $w_k$; and $f(x_i;w_k)[-1]$ is the output of the last layer, which is the same as the output of the $softmax$ layer in the neural network. Fig.~\ref{fig_relations} illustrates a relationship between logit $p_i$ and the predicted label $\Tilde{y_i}$.
\begin{figure}[htbp]
\centering
    \subfigure{
    \begin{minipage}{8cm}
        \centering
        \addtocounter{figure}{-1}
        \includegraphics[width=8cm]{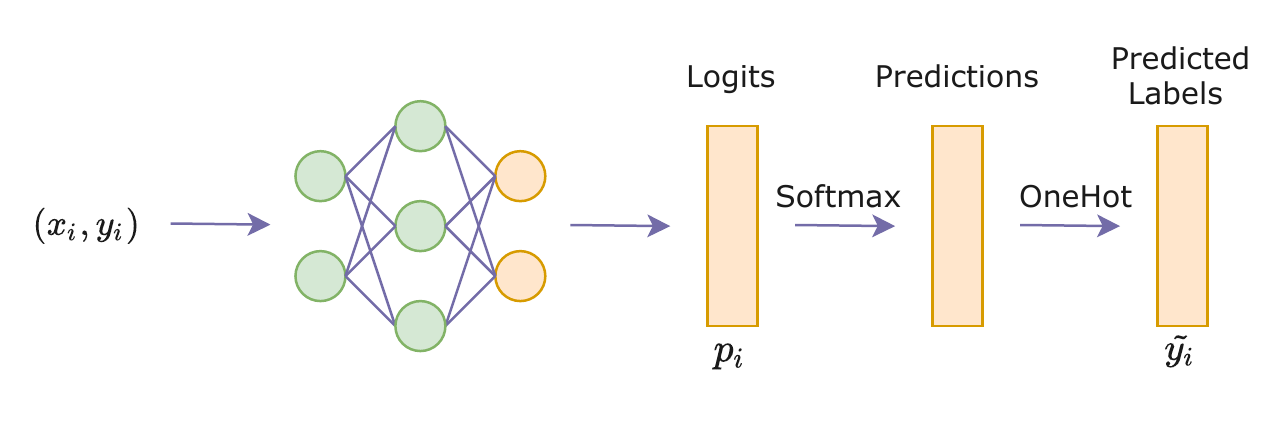}
        \caption{The relation between $p_i$ and $\Tilde{y_i}$.}
        \label{fig_relations}
    \end{minipage}
    }
\end{figure}

\subsubsection{The process in clients}

We describe how to manage logits in clients based on their private dataset here. When client $k$ is training its model, it randomly selects a batch of data from its private dataset, i.e. $(x_j, y_j)\in \Tilde{D_k}$, where $\Tilde{D_k}\subset D_K$. In local private data optimization, client $k$ collects logits for the class label $y_i$ from $\Tilde{D_K}$, i.e. $ \forall (x_j, y_j)\in \Tilde{D_k}, y_j = y_i$ as follows,
\begin{equation}
\label{collect_logit}
\begin{aligned}
& p_j = f(x_j;w_k)[-2],  \\
& p_{k,y_i} = \sum_{\forall (x_j,y_j)\in \Tilde{D_{k,y_i}}} p_j, \\
& V_{k,y_i} = V_{k,y_i} + |\Tilde{D_{k,y_i}}|,
\end{aligned}
\end{equation}

where $p_{k,y_i}$ is the sum of logits which belong to class $y_i$ in client $k$, $\Tilde{D_{k,y_i}}$ contains all instances with label $y_i$ in $\Tilde{D_k}$, and $|\Tilde{D_{k,y_i}}|$ is the number of instances with class label $y_i$ in $\Tilde{D_k}$. $V_{k,y_i}$ counts the number of logits in class $y_i$ from a randomly selected dataset $\Tilde{D_k}$. When client $k$ finishes its training, the logit is updated as follows,
\begin{equation}
\label{avg_logit}
\begin{aligned}
& p_{k,y_i} = \frac{p_{k,y_i}}{V_{k,y_i}+1},
\end{aligned}
\end{equation}
where $p_{k,y_i}$ is an average logits for class $y_i$. Note that we add one in the denominator to avoid an extreme situation with $V_{k,y_i}=0$. Eq. (\ref{collect_logit}) and (\ref{avg_logit}) explain how to handle logits in the client. After computing $p_{k,y_i}$ for all the class labels, client $k$  transmits the aggregated logits in the format of $(p_{k,y_i}, y_i)$ to the server.

\subsubsection{The process in the server}
We explain how to process the logits in the server here. Given that the server may receive the logits from the clients at different time stamps, the collection of logits is designed to be an asynchronous process. Thus, the server has to maintain a storage for storing the logits and their corresponding labels,  i.e. $(p_{k,y_i}, y_i)$, from the clients. Let $D_l$ be the logits received by the server from the clients. We define $D_{l,y_i}$ as the logits in $D_l$ with label $y_i$, i.e. $D_l = D_{l,y_1} \cup D_{l,y_2} \cup ... \cup D_{l,y_c}$, if there are $c$ classes. When the server receives logits from a client, $D_l$ will be updated to include the newly received logits as follows,
\begin{equation}
\label{server_collect_logits}
\begin{aligned}
& D_{l,y_i} \leftarrow  D_{l,y_i} \cup p_{k,y_i},
\end{aligned}
\end{equation}
where $D_{l,y_i}$ is the dataset storing logits in class $y_i$ from the clients. Note that the server only stores the aggregated logits according to their classes. When a client requests for logits to train a model, the server will compute the averaged the logits as follows,
\begin{equation}
\label{server_avg_logits}
\begin{aligned}
& p_{s,y_i} = \frac{1}{|D_{l,y_i}|} \sum_{j=1}^{|D_{l,y_i}|} p_{k,y_j},\forall p_{k,y_j}\in D_{l,y_i},
\end{aligned}
\end{equation}
where $|D_{l,y_i}|$ is the number of logits stored in $D_{l,y_i}$ for class $y_i$. Eq. (\ref{server_avg_logits}) is computed by averaging all the logits stored in $D_{l,y_i}$ for class $y_i$. Using this method, the server can compute the average logits for all the classes. The average logits of all the classes, $(p_{s,y_i}, y_i), \forall y_i$, will be transmitted from the server to all the clients to enrich their knowledge in training. The number of average logtis is the same as the number of classes.

 The averaging step mixes the logits from different clients, but in fact it captures important features of the instances from a certain class among the clients. The information captured by the average logits can bring lots of additional knowledge to the clients \cite{hinton2015distilling}.

\subsection{Training on logits}
We have discussed how to collect logits in the clients and how to process the logits in the server so far. In the following, we will discuss how to utilize logits to update the models in the clients.

We consider that each client $k$ receives the average logits from the server. For an instance $(x_i, y_i)\in \Tilde{D_k}$, client $k$ performs local model training on both its private data and the average logits. At first, client $k$ matches the average logit for $(x_i, y_i)$ according to its label $y_i$, i.e., the training instance becomes $(x_i, p_{s, y_i}, y_i)$. The objective function for logit optimization is
\begin{equation}
\label{1_logit_optimization}
\begin{aligned}
 p_i = f(x_i;w_k)[-2],
 \end{aligned}
\end{equation}
\begin{equation}
\label{2_logit_optimization}
\begin{aligned}
 \min_{w_k}\ \ L_{logit}=\frac{1}{|\Tilde{D_k}|}\sum_{i=1}^{|\Tilde{D_k}|}{l_{logit}(p_i, p_{s, y_i})},
\end{aligned}
\end{equation}
where $p_i$ is the logit of $x_i$ in model $w_k$, and $l_{logit}$ computes the loss between logit $p_i$ and the average logit of class $y_i$ from the server $p_{s,y_i}$. The goal of this objective function is for the client $k$ to learn additional knowledge as side information from the private data of other clients through the average logits.

The loss function of client $k$ is composed of both private data training and logit optimization as below.
\begin{equation}
\label{sum_loss_func}
\begin{aligned}
 \min_{w_k}\ \ L_k=\frac{1}{|\Tilde{D_k}|}\sum_{i=1}^{|\Tilde{D_k}|}{(l_{private}(\Tilde{y_i}, y_i)} \\
 + \alpha l_{logit}(p_i, p_{s, y_i})),
\end{aligned}
\end{equation}
where $\alpha$ controls the trade-off between $l_{private}$ and $l_{logit}$. This parameter is set to one in our experiments.
The loss function $L_k$ can be applied to solve the heterogeneous FL problem in Eq. (\ref{Hetero_FL_global_objective}). 

The whole process of our method is described in Algorithm \ref{Algorithm_FedHe}. At the beginning of the training process, both the server and the clients do not have any logits. In order to solve the cold-start problem, the clients begin with  local training first and optimize their models using Eq. (\ref{local_private_data_objective}). The clients continue to collect their logits and transmit these logits to the server. When the server receives logits from the clients, the FedHe training process will begin to incorporate average logits. 

\begin{algorithm}[htbp]
    \caption{FedHe}
    \label{Algorithm_FedHe}
    \KwIn {Private dataset $D_k, k\in\{1,...,K\}$, $K$ clients and their weights $w_1, ..., w_K$.}
    \KwOut {Optimal weights for all clients $w_1, ..., w_K$.}
    \textbf{Server process:}\\
    \For {each round $i=1,2,...$ do}
    {
        Receive any logits from one client.\\
        Gather logits in $D_l$ as (\ref{server_collect_logits}).\\
        Compute the average logits $p_{s,y_i}$ by (\ref{server_avg_logits}).\\
        Transmit the average logits $(p_{s,y_i}, y_i), \forall y_i$ to the client requesting for updating.
    }
    \textbf{Client processes:} \\
    \While {random clients $k, k\in {1,...,K}$}
    {
        Receive average logits $(p_{s,y_i}, y_i), \forall y_i$ from a server.\\
        \For {each batch $\Tilde{D_k} \in D_k$}
        {
            \For {each instances $(x_i, y_i)\in \Tilde{D_k}$}
            {
                Match an average logits with the same label $(p_{s,y_i}, y_i)$ to the instance $(x_i, y_i)$. \\
                Update $w_k$ by \ref{sum_loss_func}.
            }
            Collect logits $p_{k, y_i}, \forall y_i$ by \ref{collect_logit}.
        }
        Update the logits $p_{k, y_i}, \forall y_i$ by \ref{avg_logit}.\\
        Transmit updated logits $p_{k, y_i}, \forall y_i$ to the server.
    }
\end{algorithm}

\begin{table*}[htbp]
    \centering
    \begin{tabular}{c|c|c|c|c}
      Model & 1st-CNN filters & 2nd-CNN filters & 3rd-CNN filters & dropout rate \\
      \hline
      \hline
      0  & 128 & 256 & / & 0.2 \\
      1  & 128 & 384 & / & 0.2 \\
      2  & 128 & 512 & / & 0.2 \\
      3  & 256 & 256 & / & 0.3 \\
      4  & 256 & 512 & / & 0.4 \\
      5  & 64 & 128 & 256 & 0.2 \\
      6  & 64 & 128 & 192 & 0.2 \\
      7  & 128 & 192 & 256 & 0.2 \\
      8  & 128 & 128 & 128 & 0.3 \\
      9  & 128 & 128 & 198 & 0.3 \\
    \end{tabular}
    \caption{Heterogeneous Model Architectures}
    \label{table_model_archi}
\end{table*}

\begin{table*}[htbp]
    \centering
    \begin{tabular}{c|ccccc}
       \multirow{2}*{Model} & \multicolumn{5}{c}{Communication Overheads in each rounds for one client} \\
      \cline{2-6}
      ~ & Logits & Model weights & Transmitting samples & Total & Reduced rate (n=10) \\
      \hline
      \hline
      FedAvg  & N/A & 324,672 & N/A & 324,672 & 0\\
      \textbf{FedHe}  & $10\times 11$ & N/A & N/A & 110 & $>99.9\%$ \\
      FedMD  & $n\times 10$ & N/A & $n\times784$ & $n\times794$ & $97.6\%$
    \end{tabular}
    \caption{The details of communication overheads for MNIST}
    \label{table_communication_overheads_MNIST}
\end{table*}

\begin{table*}[htbp]
    \centering
    \begin{tabular}{c|ccccc}
       \multirow{2}*{Model} & \multicolumn{5}{c}{Communication Overheads in each rounds for one client} \\
      \cline{2-6}
      ~ & Logits & Model weights & Transmitting samples & Total & Reduced rate (n=10) \\
      \hline
      \hline
      FedAvg  & N/A & 326,976 & N/A & 326,976 & 0\\
      \textbf{FedHe}  & $10\times11$ & N/A & N/A & 110 & $>99.9\%$ \\
      FedMD  & $n\times10$ & N/A & $n\times3072$ & $n\times3082$ & $90.6\%$
    \end{tabular}
    \caption{The details of communication overheads for CIFAR-10}
    \label{table_communication_overheads_CIFAR10}
\end{table*}

\section{Experiments}
To illustrate the performance of FedHe, we conduct experiments and analyse their result in this section. Our experiments compare the model accuracy of FedHe with the state-of-the-art methods to demonstrate the power of FedHe in training heterogeneous models. Moreover, we analyse the communication overheads of FedHe to show its efficiency compared with the state-of-the-art methods.

\subsection{Experiment setting}
We conduct the experiments on the MNIST and CIFAR-10 datasets. The MNIST dataset includes 60000 28$\times$ 28 handwriting images from  one to ten, while the CIFAR-10 \cite{krizhevsky2009learning} contains 60000 32$\times$ 32 images in ten different classes.
FedHe can support both homogeneous FL and heterogeneous FL. We conduct the experiments in both settings. The first setting requires all the clients to keep a homogeneous model. The second setting allows all the clients to have different model architectures. The model architectures in heterogeneous FL is shown in  Table \ref{table_model_archi}, in which the last layer is the dense layer. We select model 9 in Table \ref{table_model_archi} to be the model in homogeneous FL. These model architectures are the same as \cite{li2019fedmd}. We have ten clients in the experiments. The learning rate is 0.001 and the trade-off parameter $\alpha$ is set to 1. The inner training epoch in clients is 3, i.e., a client randomly selects 3 batches of dataset to train its model, which is appropriate to balance between the model accuracy and the limitation of resources in the edge devices. The images are subtracted by the mean of the whole training dataset. Other than that, the data are not preprocessed before training.

We compare FedHe with three baselines. Different baselines are utilized in different experiment settings.

\begin{itemize}
    \item \textbf{FedAvg}\cite{mcmahan2017communication}: A basic model in FL, which aggregates model weights in the server. It can support only homogeneous model training, so it is used only in the homogeneous FL experiments.
    \item \textbf{FedMD}\cite{li2019fedmd}: This model applies logits from a large public dataset for knowledge distillation. This model sets 10\% of the training dataset to be the public dataset. This baseline is deployed in both homogeneous FL and heterogeneous FL experiments.
    \item \textbf{Private}: All clients train individual models only with own private dataset and do not communicate. We use this baseline to studys the learning efficiency of FedHe. This baseline is deployed in both homogeneous and heterogeneous FL experiments.
\end{itemize}

In the following, we will first analyze the communicate efficiency of FedHe compared with the baseline methods in our experiment settings. Then, we will conduct homogeneous and heterogeneous FL experiments on MNIST and CIFAR-10 to show the overall model accuracy. We will further evaluate the model accuracy of each client in heterogeneous FL to show how FedHe improves the capabilities of the clients.

\begin{figure*}[htbp]
\centering
    \subfigure{
    \begin{minipage}{5cm}
        \centering
        \includegraphics[width=5cm]{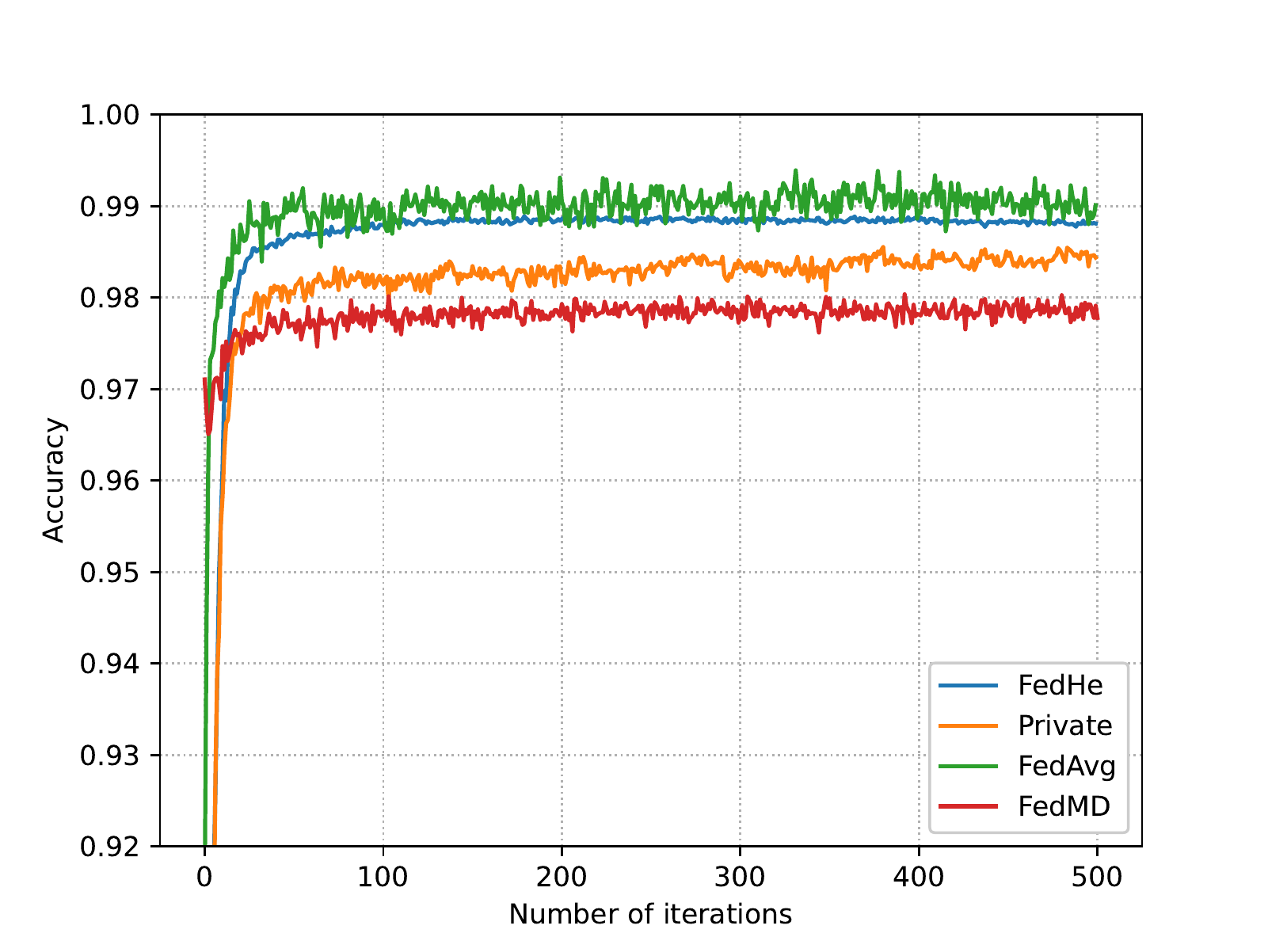}
        \caption{Accuracy of MNIST in homogeneous FL}
        \label{fig_acc_homo_mnist}
    \end{minipage}
    }
    \subfigure{
    \begin{minipage}{5cm}
        \centering
        \includegraphics[width=5cm]{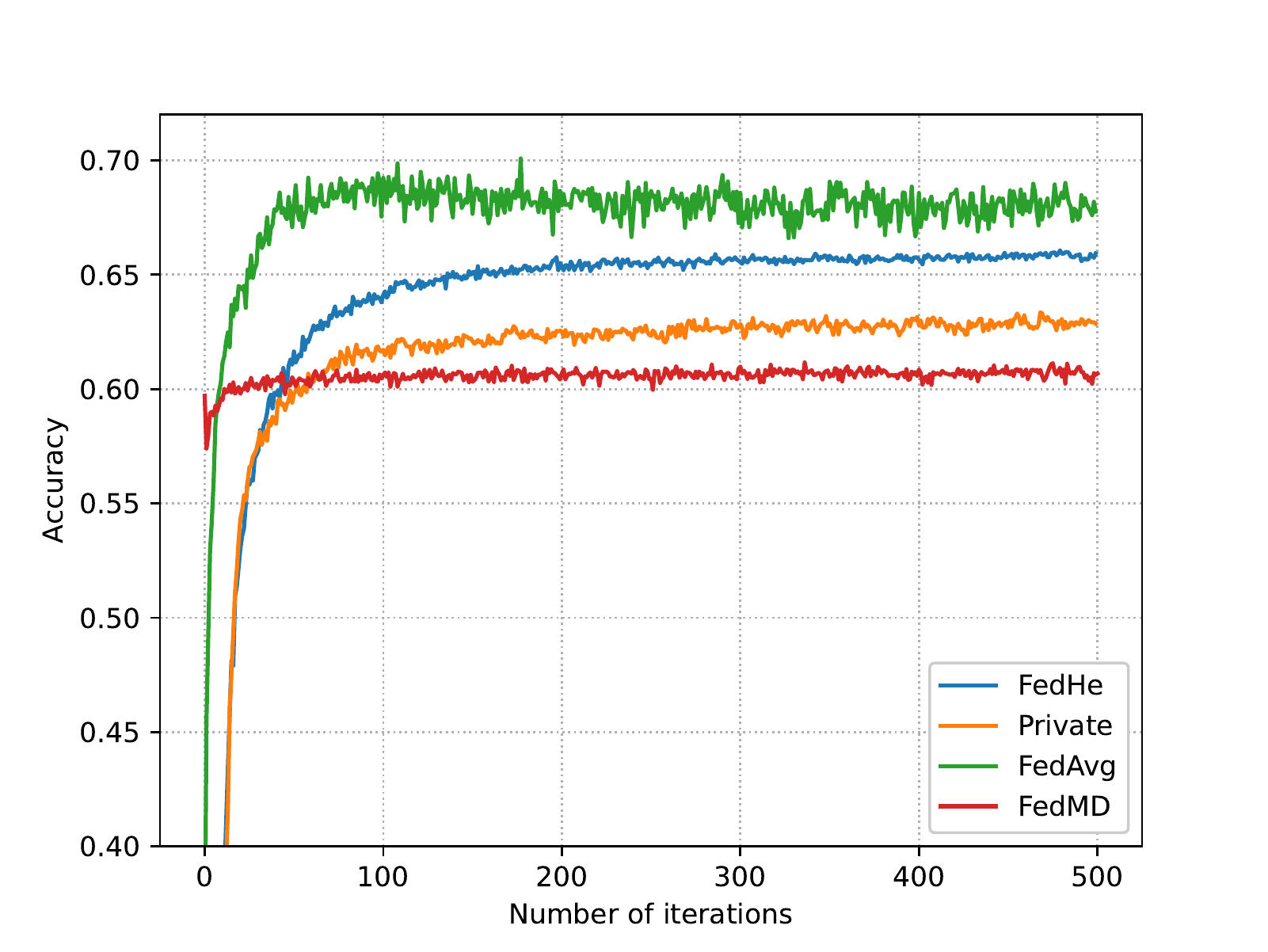}
        \caption{Accuracy of CIFAR10 in homogeneous FL}
        \label{fig_acc_homo_cifar10}
    \end{minipage}
    }
    \subfigure{
    \begin{minipage}{5cm}
        \centering
        \includegraphics[width=5cm]{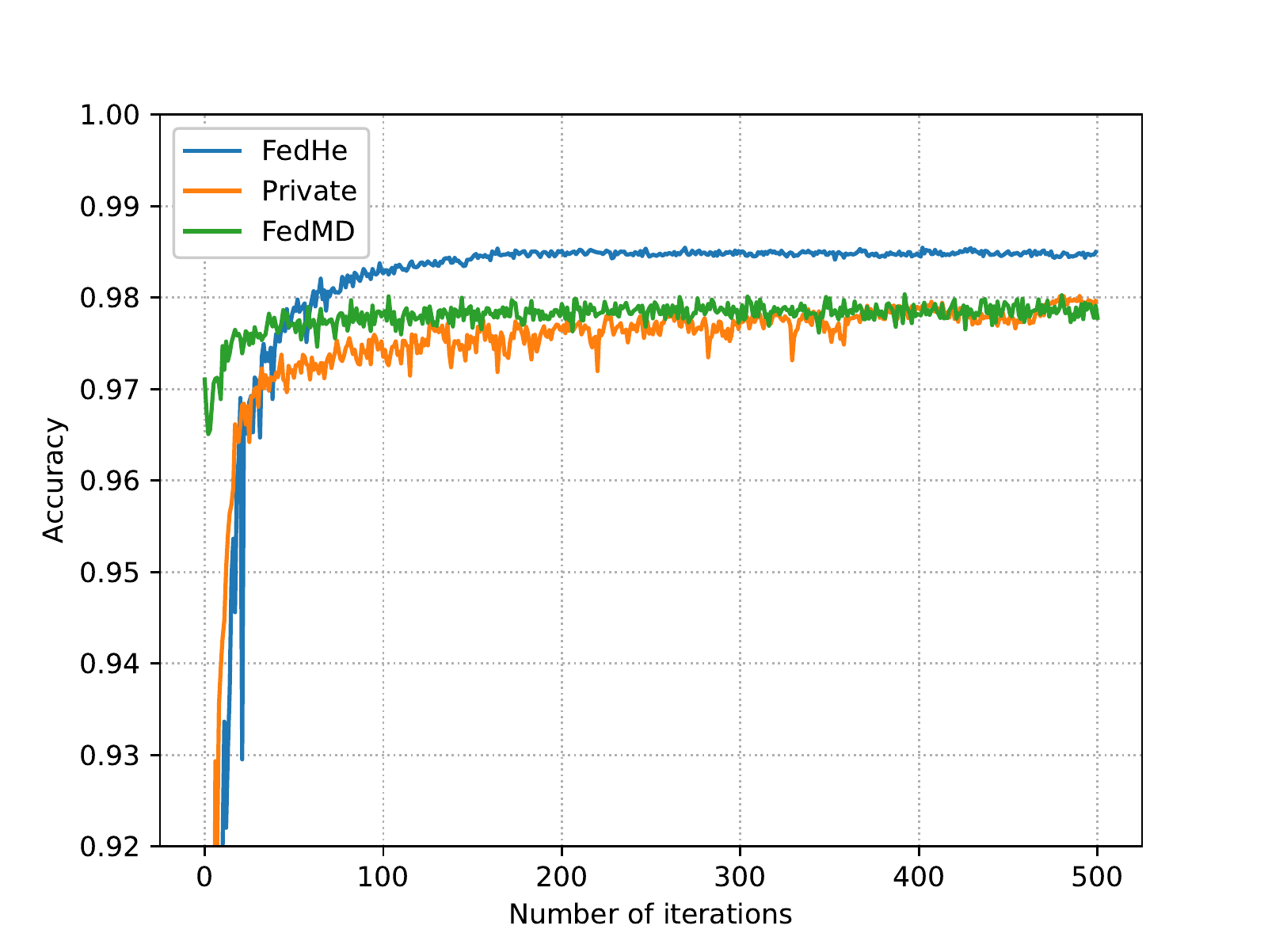}
        \caption{Accuracy of MNIST in heterogeneous FL}
        \label{fig_acc_hete_mnist}
    \end{minipage}
    }

    \subfigure{
    \begin{minipage}{5cm}
        \centering
        \includegraphics[width=5cm]{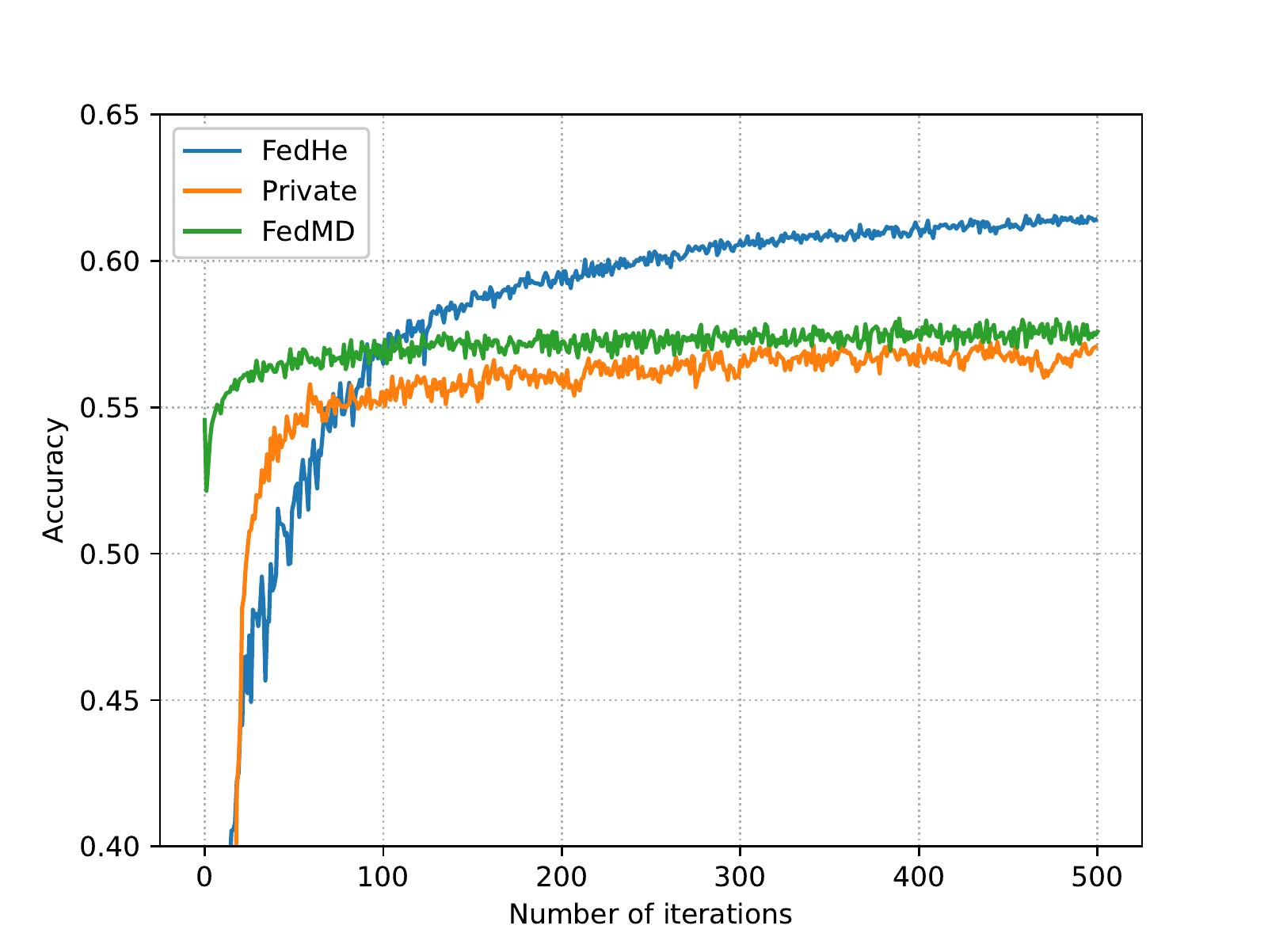}
        \caption{Accuracy of CIFAR10 in heterogeneous FL}
        \label{fig_acc_hete_cifar10}
    \end{minipage}
    }
    \subfigure{
    \begin{minipage}{5cm}
        \centering
        \includegraphics[width=5cm]{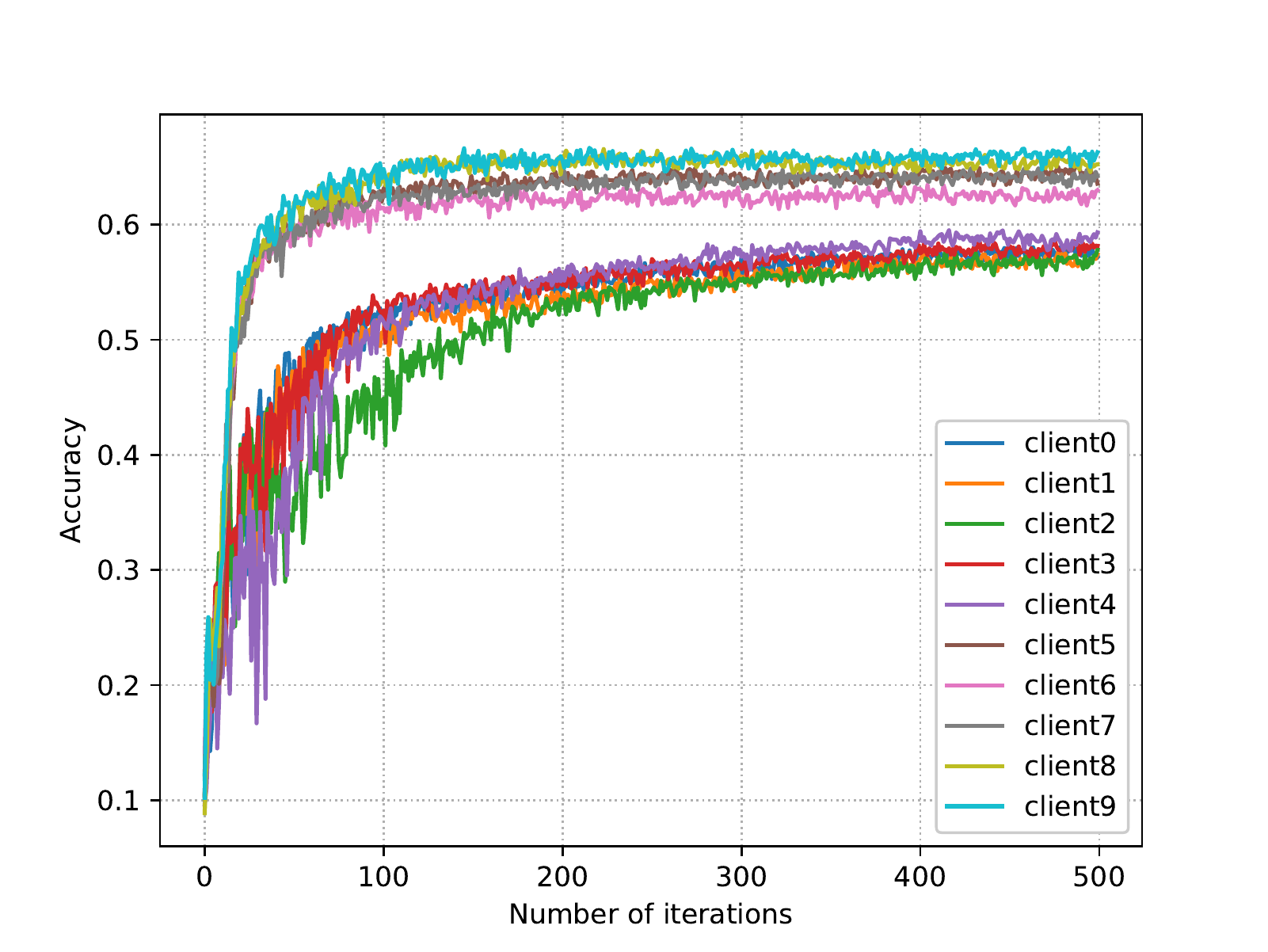}
        \caption{FedHe: Accuracy of each clients in CIFAR-10}
        \label{fig_acc_individual}
    \end{minipage}
    }
    \subfigure{
    \begin{minipage}{5cm}
        \centering
        \includegraphics[width=5cm]{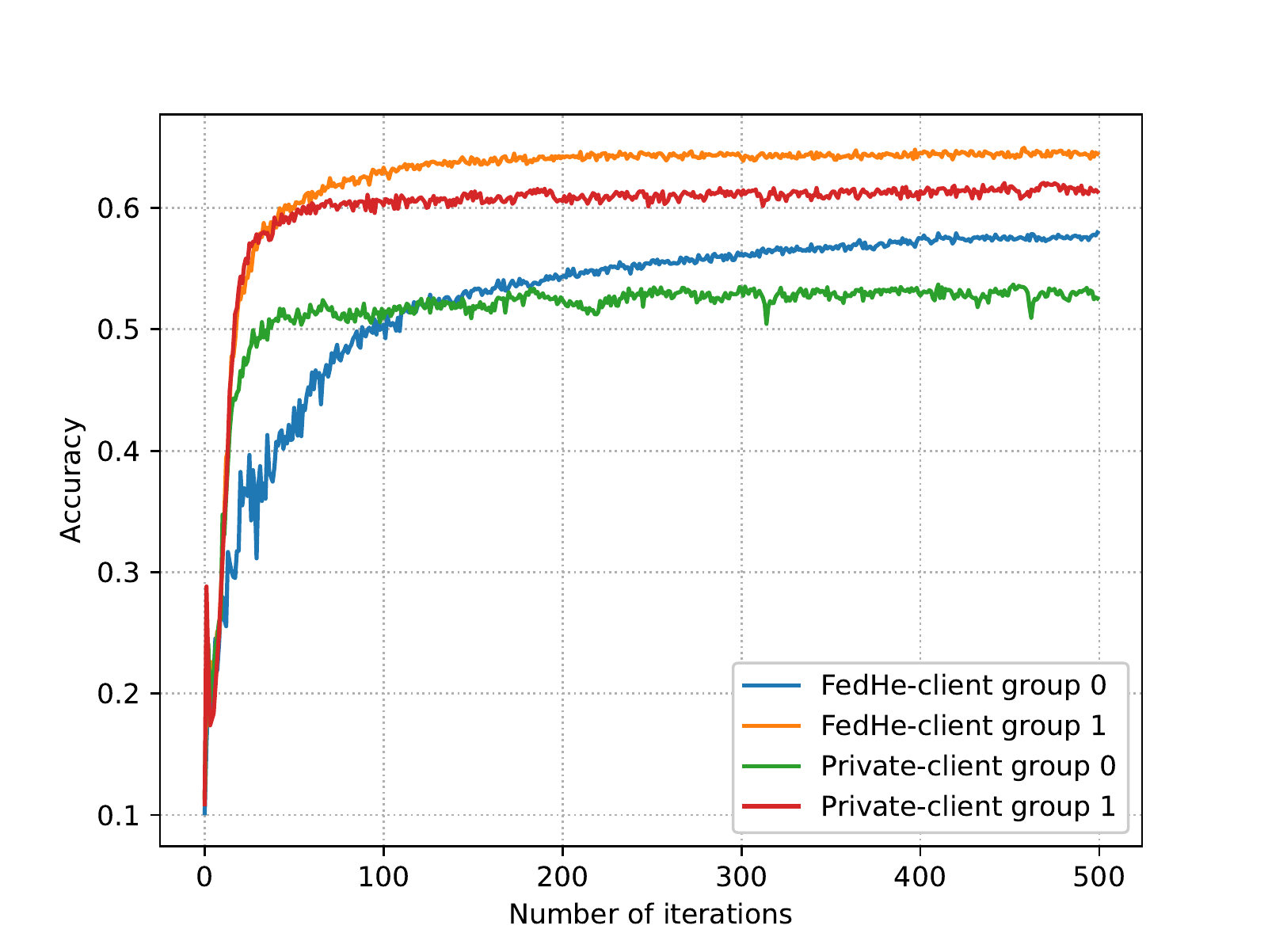}
        \caption{Accuracy of each client groups in CIFAR-10}
        \label{fig_acc_compared_individual}
    \end{minipage}
    }
\end{figure*}


\subsection{Analysis on communication overheads}
We compare the communication overheads of FedHe with FedAvg and FedMD. Table~\ref{table_communication_overheads_MNIST} and Table~\ref{table_communication_overheads_CIFAR10} show the details of communication overheads in our experiments. The model weights come from the neural networks trained in the clients. The logits are inputs of the softmax layer, which are deployed in FedHe and FedMD. Transmitting samples are the images used from the public dataset in FedMD. The clients are required to obtain those data samples from the public dataset.
Since the dimensions of images are different between MNIST and CIFAR-10, we use two tables to show their results.

From Table~\ref{table_communication_overheads_MNIST}, the number of weights in model 9 is 324,672, so the communication cost for FedAvg is 324,672. A logit is represented by $(p_i, y_i)$. The dimension of $p_i$ is 10, given 10 classes in MNIST and the dimension of $y_i$ is 1. Therefore, the dimension of one logit is 11. Each client transmits its logit information $(p_{k,y_i}, y_i)$ of all the ten classes to the server, so the total communication overheads are $11\times 10$ in FedHe. For FedMD, each client transmits both the logits and images from the public dataset. The dimension of an image in MNIST is $28\times28=784$. Therefore, the cost for transmitting samples in FedMD is $n\times 794$, where $n$ is the number of images.
Similarly, we show the communication overheads for CIFAR-10 in Table~\ref{table_communication_overheads_CIFAR10}. We also compute the reduced rates on communication overheads taking FedAvg as a baseline in Table~\ref{table_communication_overheads_MNIST} and Table~\ref{table_communication_overheads_CIFAR10}.

From the results, we see that FedHe can reduce communication overheads significantly by exchanging only logits, instead of model parameters. Even if we set a small $n$=10 for FedMD in our experiments, FedHe still reduced more communication overheads as it does not need to exchange images as in FedMD. FedHe also  does not require a common public dataset and better preserve data privacy. Note that the communication overheads are not the same for different datasets in FedMD as it depends on the size of the images. In addition, the communication overheads remain the same in both homogeneous FL and heterogeneous FL for FedHe and FedMD as they are calculated for each client per round.


\subsection{Analysis of homogeneous FL}
We conduct two experiments in homogeneous FL on MNIST and CIFAR-10, respectively. We present their results in Fig.~\ref{fig_acc_homo_mnist} and Fig.~\ref{fig_acc_homo_cifar10}. The highest model accuracy is achieved by FedAvg, which reaches 99\% in MNIST and almost 69\% in CIFAR-10. We also observe that  FedAvg converges faster than the other methods. This observation is reasonable because FedAvg transmits model weights in model updates. However, when we consider the communication cost, we can see that FedHe is much more efficient as shown in Table~\ref{table_communication_overheads_MNIST} and Table~\ref{table_communication_overheads_CIFAR10}.


Since the clients transmit only logit information to the server in FedHe, its communication overheads are less than 0.1\% of that in FedAvg. Although its communication overheads are so small, FedHe still obtains very high accuracy compared with FedAvg. FedHe achieves model accuracy of 98.9\% in MNIST and 66\% in CIFAR-10. The baseline Private method reaches 98.4\% in MNIST and 62.5\% in CIFAR-10, which is higher than another baseline FedMD that obtained 97.9\% in MNIST and 60.9\% in CIFAR-10. We do not show the pretraining process of FedMD in Fig.~\ref{fig_acc_homo_mnist} and Fig.~\ref{fig_acc_homo_cifar10}, which is for training on their public datasets until convergence. Due to  pretraining, the results of FedMD is higher than all the other methods at the beginning of the training.

We observe that the results of FedHe are better than FedMD, since the logits in FedHe are aggregated and shared with the clients by the server, i.e., the knowledge from different clients are collected and exchanged through the server. For FedMD, its logits are built locally based on a subset of the public dataset. If this subset of public dataset does not have sufficient knowledge, the clients may find it difficult to learn more information than their own private datasets.

\subsection{Analysis of heterogeneous FL}
We conduct the experiments for heterogeneous FL on  MNIST and CIFAR-10 in Fig.~\ref{fig_acc_hete_mnist} and Fig.~\ref{fig_acc_hete_cifar10}. In heterogeneous FL, all the clients may keep different architectures of their neural networks. Since FedAvg does not support heterogeneous models, it can not be used as a baseline in heterogeneous FL. From the experiment results, FedHe achieves the highest model accuracy, 98.5\% in MNIST and 62\% in CIFAR-10.
Private obtains lower accuracy than FedHe, which indicates that exchange of logits in FedHe is valuable. We also discover that FedMD performs better than Private in these experiments.
Private attains 98\% in MNIST and 57\% in CIFAR-10, while FedMD achieves 98\% in MNIST and 57.5\% in CIFAR-10. These results indicate that FedMD can support heterogeneous models, though it still has a large performance gap with FedHe. The clients transmit only small amount of information for knowledge exchange in FedHe, but it achieves superior performance. Compared with FedMD, FedHe also consumes less bandwidth according to Table~\ref{table_communication_overheads_MNIST} and Table~\ref{table_communication_overheads_CIFAR10}.

In these experiments, we find that the convergence speed of FedHe is a little bit slower than Private in Fig.~\ref{fig_acc_hete_cifar10}. It may be due to knowledge communication in the training process. The clients have to learn from both their private dataset and the average logits from the server. The two components may lead to different optimized directions in the training process. Nevertheless, the knowledge from the average logits will still improve the model accuracy eventually. Although the convergence speed of FedHe is slower than Private, its model accuracy is much higher than Private.

We do not compare the convergence speed between FedMD and FedHe, since FedMD has a pretraining scheme. It is unfair to compare two methods, one with pretraining and one without. In contrast, we can compare the convergence speed between FedHe and FedAvg because they have similar training processes. In both methods, the clients train their models first, then transmit their knowledge to the server. The server aggregates the knowledge and sends back to the clients.


\subsection{Performance of individual clients in FedHe}
To illustrate the efficiency of FedHe and show how logits improve the capabilities of clients, we show the model accuracy of different clients in Fig.~\ref{fig_acc_individual} and Fig.~\ref{fig_acc_compared_individual}.

Fig.~\ref{fig_acc_individual} shows the  model accuracy of individual clients from FedHe on CIFAR-10 with the same setup as in Fig.~\ref{fig_acc_hete_cifar10}. We discover that clients 0 to 4 obtain lower model accuracy than clients 5 to 9, which is an interesting result. Clients 0 to 4 achieve 57-59\% in model accuracy, while clients 5 to 9 obtain near 62-66\%. The client models are shown in Table~\ref{table_model_archi}, from which we know that the first five models are 2-CNNs and the last five models are 3-CNNs. Our results show that heavier models achieve higher model accuracy, which is common in machine learning. Our experiment demonstrates that FedHe can support training of heterogeneous models in the clients efficiently. It is possible for a client to choose an affordable model architecture that fits its needs, while still benefits from knowledge exchange with the server and other clients.

Therefore, if clients keep heavier models in these experiments, the accuracy will be better than the results shown in the paper. Moreover, this analysis also proves that the dissatisfactory accuracy comes from the model architectures, not because of FedHe.

In Fig.~\ref{fig_acc_compared_individual}, we divides ten clients into two groups. The first group called client group 0, is comprised of client 0 to client 4, which are 2-CNNs models. The second group called client group 1, is consisted of client 5 to 9, i.e., 3-CNNs models. The average model accuracy of the two client groups for FedHe and Private are shown in Fig.~\ref{fig_acc_compared_individual}. It shares the same experiment setup as in Fig.~\ref{fig_acc_hete_cifar10}.

We find that the model accuracy of both groups in FedHe are better than in Private. Client group 0 from FedHe reaches 58\%, while the group 0 from Private only achieves 52-53\%. Client group 1 in FedHe achieves 65\%, while client group 1 in Private obtains 61\%. Based on these results, we find that logits not only improve the performance of lighter clients. In fact, the clients with heavier models also gain a lot of knowledge from the logits.
Our FedHe method successfully boosts the performance of clients by exchanging logits between the server and the clients with only small communication overheads.


\section{Conclusions}

In this paper, we proposed a novel FL scheme, called FedHe, which supports training of heterogeneous models in clients by exchanging logits. Logits capture important features of instances in individual classes. The logits are aggregated at the server and shared with the clients as side information to enhance model training.
Our solution supports heterogeneous models and asynchronous communications. The clients transmit logit information, instead of model parameters, for knowledge exchange.
The communication overhead of FedHe is less than one percent of traditional FedAvg. In addition, FedHe does not require any public dataset for pretraining, so it can preserve data privacy. Our experimental results demonstrated that FedHe can train heterogeneous models with satisfactory model accuracy and significantly reduced communication overheads.

\section*{Acknowledgment}
We would like to acknolwedge the RGC Grant Resaerch Fund no. 17203320 from Hong Kong.

\bibliographystyle{IEEEtran}
\bibliography{refer}


\end{document}